\documentclass[conference]{IEEEtran}
\IEEEoverridecommandlockouts
\usepackage{enumitem}
\usepackage{booktabs}
\usepackage[table]{xcolor} % 用于表格颜色
\usepackage{multirow}
\usepackage{booktabs}
\usepackage{multirow}
\usepackage{makecell}
\usepackage[table]{xcolor}
\usepackage{subcaption} % 用于 subfigure 环境
\usepackage{graphicx}
\usepackage{cite}
\usepackage{amsmath,amssymb,amsfonts}
\usepackage{algorithmic}
\usepackage{graphicx}
\usepackage{textcomp}
\usepackage{xcolor}
\makeatletter
\def\IEEEauthorrefmark#1{\textsuperscript{#1}} 
\makeatother

\def\BibTeX{{\rm B\kern-.05em{\sc i\kern-.025em b}\kern-.08em
    T\kern-.1667em\lower.7ex\hbox{E}\kern-.125emX}}
\begin{document}

\title{Disentangling Fact from Sentiment: A Dynamic Conflict-Consensus Framework for Multimodal Fake News Detection}

\author{
\IEEEauthorblockN{
Weilin Zhou\IEEEauthorrefmark{1}\IEEEauthorrefmark{2}$^{\star}$,\;
Zonghao Ying\IEEEauthorrefmark{3}$^{\star}$,\;
Rongchen Zhao\IEEEauthorrefmark{4},\;
Chunlei Meng\IEEEauthorrefmark{5},\;
Quanchen Zou\IEEEauthorrefmark{2}$^{\dagger}$,\\
Deyue Zhang\IEEEauthorrefmark{2},\;
Enhao Gu\IEEEauthorrefmark{1},\;
Mingze Liu\IEEEauthorrefmark{6},\;
Dongdong Yang\IEEEauthorrefmark{2},\;
Xiangzheng Zhang\IEEEauthorrefmark{2}
}\\
\IEEEauthorblockA{\IEEEauthorrefmark{1}Xinjiang University}
\IEEEauthorblockA{\IEEEauthorrefmark{2}360 AI Security Lab}
\IEEEauthorblockA{\IEEEauthorrefmark{3}Beihang University}
\IEEEauthorblockA{\IEEEauthorrefmark{4}South China University of Technology}
\IEEEauthorblockA{\IEEEauthorrefmark{5}Fudan University}
\IEEEauthorblockA{\IEEEauthorrefmark{6}Shenzhen University}
}
\maketitle

\begingroup
\renewcommand\thefootnote{}
\footnote{* Equal contribution.}
\footnote{$\dagger$ Corresponding author: Quanchen Zou.}
\footnote{The work was done at 360 AI Security Lab.}
\addtocounter{footnote}{-3}
\endgroup

\begin{abstract}
Prevalent multimodal fake news detection relies on consistency-based fusion, yet this paradigm fundamentally misinterprets critical cross-modal discrepancies as noise, leading to over-smoothing, which dilutes critical evidence of fabrication. Mainstream consistency-based fusion inherently minimizes feature discrepancies to align modalities, yet this approach fundamentally fails because it inadvertently smoothes out the subtle cross-modal contradictions that serve as the primary evidence of fabrication. To address this, we propose the Dynamic Conflict-Consensus Framework (DCCF), an inconsistency-seeking paradigm designed to amplify rather than suppress contradictions. First, DCCF decouples inputs into independent Fact and Sentiment spaces to distinguish objective mismatches from emotional dissonance. Second, we employ physics-inspired feature dynamics to iteratively polarize these representations, actively extracting maximally informative conflicts. Finally, a conflict-consensus mechanism standardizes these local discrepancies against the global context for robust deliberative judgment.Extensive experiments conducted on three real world datasets demonstrate that DCCF consistently outperforms state-of-the-art baselines, achieving an average accuracy improvement of 3.52\% .
\end{abstract}


\begin{figure}[htbp]
    \centering
    \begin{subfigure}{\columnwidth}
        \centering
        \includegraphics[width=1.0\columnwidth]{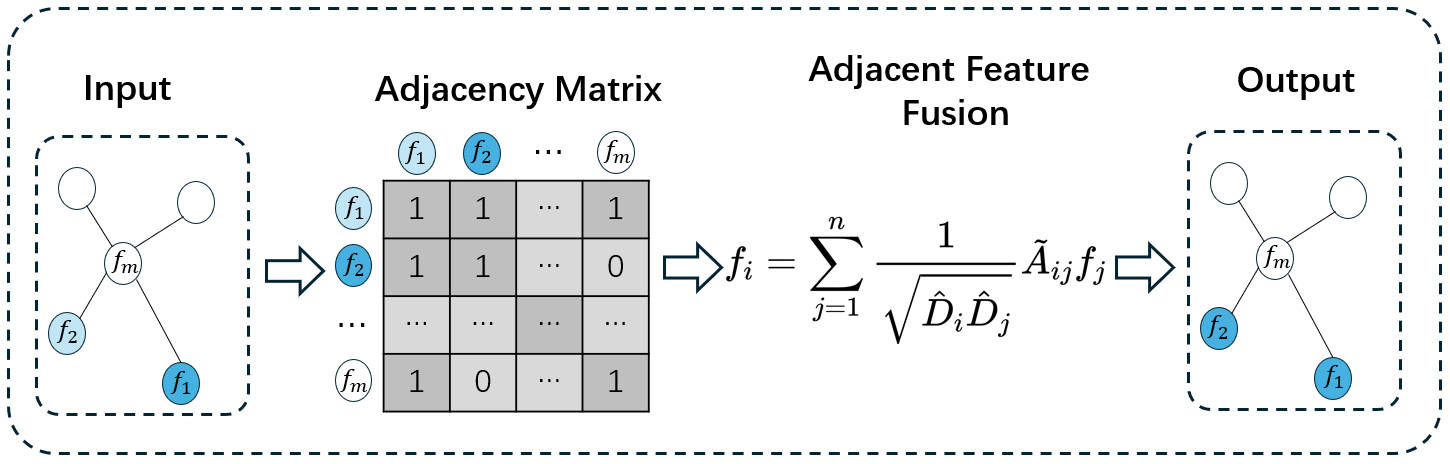}
        \caption{Graph-based methods}
        \label{fig:sub_distillation_and_temp}
    \end{subfigure}
    \vspace{0.15cm} % Keeping this small vspace between subfigures as it's intentional internal layout
    \begin{subfigure}{\columnwidth}
        \centering
        \includegraphics[width=1.0\columnwidth]{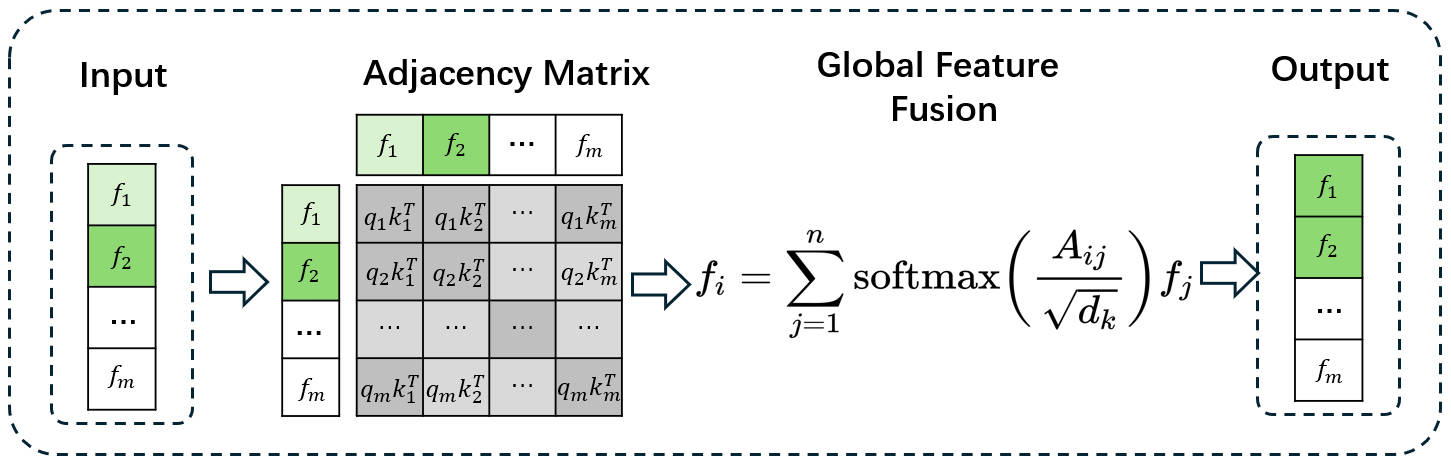}
        \caption{Attention-based method}
        \label{fig:sub_weight_and_heads}
    \end{subfigure}
    \caption{Schematic of inconsistency distortion, where $f_1$ and $f_2$ denote a pair of inconsistent features (e.g., conflicting text and image). (a) Graph-based methods blur this specific conflict by averaging neighbor features via edges. (b) Attention-based methods dilute the inconsistency by aggregating global features via weights, resulting in homogenized outputs.}
    \label{fig:framework_comparison_vertical}
    \vspace{-9pt}
\end{figure}

\section{Introduction}
Digital platforms amplify misinformation via deceptive multimodal content. While automated defense is critical, mainstream approaches predominately employ consistency-based fusion to align cross-modal features. However, we argue this premise is fundamentally flawed: the essence of fake news lies in inconsistency, subtle clashes between visual and textual evidence. By prioritizing alignment and treating discrepancies as noise, existing models inadvertently dilute the conflicting signals that serve as the primary evidence of fabrication. Consequently, effective detection demands a paradigm shift: moving from seeking consensus to explicitly modeling and amplifying inconsistency.

Fake news detection methods generally fall into three categories. Unimodal methods analyze single data streams, yet this isolation creates information islands that overlook critical cross-modal inconsistencies. Multimodal fusion approaches, ranging from early concatenation (BMR\cite{ying2023bootstrapping}) to co-attention (SEER\cite{zhu2025seer}), typically adopt consistency-seeking paradigms. These methods inadvertently smooth out vital conflict signals by treating discrepancies as noise and conflating objective facts with subjective sentiment. Recently, Large Language Models (LLMs) have been leveraged for their reasoning capabilities in approaches like INSIDE\cite{wang2025bridging} and LIFE\cite{wang2025prompt}. Nevertheless, the inherent focus of these models on semantic alignment hinders their ability to effectively capture and amplify the fine-grained inconsistent evidence essential for robust detection.

Despite innovations, existing methods suffer from fundamental limitations: (1) Semantic Entanglement, where objective content and subjective emotion are treated as a mixed signal, blurring the distinction between what is depicted and how it is described, making it difficult to distinguish factual mismatches from emotional dissonance. We resolve this by explicitly disentangling inputs via multi-task supervision. (2) Inconsistency Attenuation, where existing approaches prioritize alignment, inadvertently treating meaningful contradictions as noise and filtering them out, effectively smoothing out the critical discrepancy signals. In contrast, our framework pioneers an inconsistency-seeking paradigm, deploying a tension field network to explicitly amplify feature repulsions and extract maximally informative conflicts as primary evidence.

To address these limitations, we propose the \textbf{D}ynamic \textbf{C}onflict-\textbf{C}onsensus \textbf{F}ramework (DCCF). Inspired by physical field theory where tension \cite{steigmann1990tension} reflects the intensity of differences, our framework adopts a novel approach that actively searches for inconsistencies. It first separates inputs into factual content and emotional tone using multi-task supervision, guided by YOLO \cite{redmon2016you} and SenticNet \cite{cambria2022senticnet}, to distinguish objective entities from subjective feelings. A fact sentiment tension field network, modeling dynamic forces between features, then iteratively refines these features to highlight their differences. This process amplifies the contrast to extract the most significant conflicts as primary evidence, while simultaneously summarizing the overall style to serve as a global reference. By evaluating specific local conflicts against this global context, DCCF achieves robust, interpretable predictions. Our main contributions are:

\begin{enumerate}
    \item We propose DCCF, a novel inconsistency-seeking paradigm for multimodal fake news detection (MFND). Unlike consistency-seeking methods that blur critical signals, our framework models feature dynamics to amplify and extract inconsistency as primary evidence.
    
    \item We introduce an end-to-end fact-sentiment tension field network that quantifies tension metrics to expose latent inconsistencies. By standardizing extreme conflicts against global consensus, it transforms abstract feature dynamics into interpretable reasoning, pinpointing the exact evidence of fabrication.
\item We validate DCCF's effectiveness through extensive experiments on widely used MFND benchmarks. Our scheme shows significant performance gains over state-of-the-art baselines, demonstrating superior reliability and robustness.
\end{enumerate}

\section{Related Work}
\subsection{Multimodal Fake News Detection}
Early unimodal methods \cite{singhal2019spotfake} were insufficient, leading to multimodal detection. Text-visual fusion often interprets images superficially. This, with isolated text features, creates information islands and weak reasoning, failing cross-modal inconsistency detection. Our multi-stage framework addresses this \cite{chen2022cross}. We extract diverse factual/sentimental features for dynamic evolution, then extract high level metrics of conflict, consensus, and inconsistency. This focus on evolved relationships, not raw features \cite{ni2021mvan}, enables robust multi-view judgment by reasoning about inconsistencies.

\subsection{Multi-Domain Fake News Detection}
Multi-domain learning models news data spanning diverse domains \cite{wang2018eann}. Approaches use hard sharing for domain-specific/cross-domain knowledge or soft sharing, like gating networks or domain memory banks. However, these methods often just adjust view weights via domain embeddings, failing to learn domain invariant/specific information \cite{zhou2023multimodal}. Concatenating embeddings may fail to account for domain dependencies. Furthermore, these are single modal, struggling with rich visual information \cite{ wang2025bridging}.

Despite their architectural variations, existing unimodal, fusion-based \cite{zhu2025seer}, and LLM-driven \cite{wang2025bridging} methods predominantly rely on a consistency-seeking paradigm \cite{wang2025prompt}, which aligns features and inadvertently smooths out the critical discrepancies indicative of deception. In contrast, our DCCF pioneers an inconsistency-seeking paradigm, explicitly modeling and amplifying these cross-modal conflicts to leverage inconsistency as the primary evidence for detection \cite{ying2023bootstrapping}.

\section{Methodology}
\label{sec:methodology}

\begin{figure*}[t]
\centering
\includegraphics[width=1.0\textwidth]{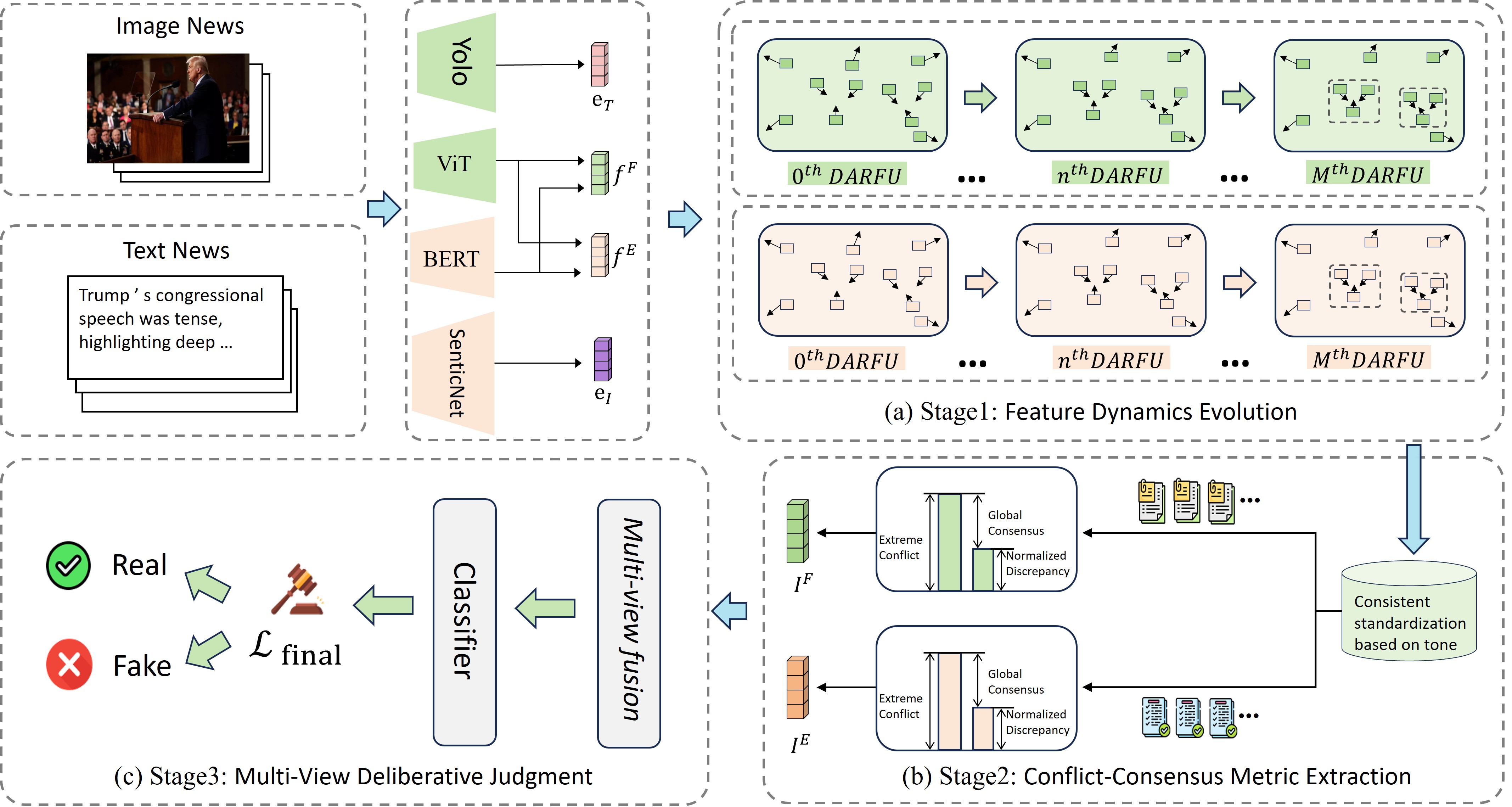}
\caption{The DCCF framework: (a) Fact-Sentiment Feature Extraction projects features into Fact ($S^F$) and Sentiment ($S^E$) spaces; (b) Feature Dynamics Evolution refines features through DARFU blocks to compute conflict/consensus; (c) Multi-View Deliberative Judgment fuses metrics for the final decision.}
\label{fig:model_architecture}
\vspace{-9pt}
\end{figure*}

We propose the DCCF framework to detect multimodal fake news by modeling feature dynamics within decoupled semantic spaces. DCCF first disentangles inputs into distinct fact and sentiment spaces via multi-task learning, then employs a tension field network to identify polarization and inconsistency. As shown in Fig. \ref{fig:model_architecture}, the framework comprises three progressive stages: (1) \textbf{Fact-Sentiment Feature Extraction}, (2) \textbf{Feature Dynamics Evolution and Conflict-Consensus Metric Extraction}, and (3) \textbf{Multi-View Deliberative Judgment}.

\subsection{Fact-Sentiment Feature Extraction}
This stage projects input text and images onto specialized fact and sentiment feature spaces. This separation effectively distinguishes objective factual inconsistencies from subjective sentimental conflicts, addressing the limitation where monolithic processing often conflates these distinct signals by anchoring features to visual objects and textual polarity, respectively.

Initial Encoding.
Using pretrained BERT \cite{devlin2019bert} and ViT \cite{dosovitskiy2021image}, we extract initial embeddings $e_T$ and $e_I$ from the raw text $T$ and image $I$:
\begin{equation}
e_T = \text{BERT}(T), \quad e_I = \text{ViT}(I)
\end{equation}

Fact Space Projection.
Two independent MLPs project $e_T$ and $e_I$ into a shared fact space, yielding $f_T^F$ and $f_I^F$. To strictly enforce objectivity and semantic alignment, we introduce an auxiliary task:
\begin{enumerate}[label=\arabic*., leftmargin=*]
\item The image $I$ is fed into a pretrained YOLO \cite{redmon2016you} to generate pseudo-labels $e_Y$ representing key factual objects (e.g., entities, locations).
\item We train the projection to accurately predict $e_Y$ via a BCE Loss $\mathcal{L}_F$, ensuring the space captures grounded reality:
\end{enumerate}
\begin{equation}
\mathcal{L}_F = \text{BCE}(\text{MLP}_{\text{classifier}}(f_I^F), e_Y)
\end{equation}

Sentiment Space Projection.
Simultaneously, two separate MLPs project $e_T$ and $e_I$ into a sentiment space, producing $f_T^E$ and $f_I^E$. We orient this space towards subjective sentiment via another auxiliary task to capture high level affective semantics:

% 使用 figure* 让宽图片跨栏显示

\begin{enumerate}[label=\arabic*., leftmargin=*]
\item Text $T$ is processed by a lexicon (e.g., SenticNet) to obtain a sentiment polarity vector $e_J$.
\item We enforce $f_T^E$ to predict this polarity using MSE Loss $\mathcal{L}_E$:
\end{enumerate}
\begin{equation}
\mathcal{L}_E = \text{MSE}(\text{MLP}_{\text{classifier}}(f_T^E), e_J)
\end{equation}

The output comprises the fact feature space $S^F = \{f_T^F, f_I^F\}$ and sentiment feature space $S^E = \{f_T^E, f_I^E\}$.

\subsection{Feature Dynamics Evolution and Conflict-Consensus Metric Extraction}
A dynamic evolution module (inspired by physical tension field theory) operates on $S^F$ and $S^E$ to amplify inconsistency (conflict) and distill global context (consensus).

Feature Dynamics Evolution.
Given input $S^{(0)} = \{f_1^{(0)}, ..., f_n^{(0)}\}$ of $n$ features, a Dynamic Feature Evolution Unit (DARFU) iterates $M$ times ($t=0$ to $M-1$):

Compute tension. Calculate tension $T_{i,j}^{(t)}$ between vectors $f_i^{(t)}, f_j^{(t)} \in S^{(t)}$, representing the potential difference in the semantic field:
\begin{equation}
T_{i,j}^{(t)} = (f_i^{(t)} - f_j^{(t)})^2
\end{equation}

Compute weights. Convert high tension to low attraction weight via softmax:
\begin{equation}
W_{i,j}^{(t)} = \text{softmax}_j(-T_{i,j}^{(t)})
\end{equation}

Aggregate and transform. Update features via residual weighted sum and non-linear transformation $g$ for state $S^{(t+1)}$:
\begin{equation}
f_i^{(t+1)} = f_i^{(t)} + g\left( \sum_{j=1}^{n} W_{i,j}^{(t)} f_j^{(t)} \right)
\end{equation}

Crucially, this iterative process acts as an inconsistency-seeking filter: it clusters semantically consistent neighbors while polarizing inconsistent ones, thereby preventing the over-smoothing of conflicting evidence common in graph-based fusion. The output is the final space $S' = S^{(M)}$ and tension matrix $T' = T^{(M-1)}$.

Conflict-Consensus Metric Extraction.
From $S'$ and $T'$, we extract two key metrics to quantify internal contradictions:

\begin{table*}[t]
    \centering
    \caption{Performance comparison of DCCF against state-of-the-art methods on Weibo, Weibo-21, and GossipCop datasets. The row “Improv." indicates the improvement of our model compared to the best-performing baseline.}
    \label{tab:main_results}
    
    % 1. 减小列间距 (默认是6pt，改成1.5pt-3pt之间)
    \setlength{\tabcolsep}{2.5pt} 
    % 2. 减小行高 (默认是1.0，改成0.85-0.9之间)
    \renewcommand{\arraystretch}{0.85}
    
    \resizebox{0.95\textwidth}{!}{% % 如果觉得太宽，可以把 \textwidth 改成 0.95\textwidth
    \begin{tabular}{lrrrrrrrrrrrr}
        \toprule
        & \multicolumn{4}{c}{\textbf{Weibo}} & \multicolumn{4}{c}{\textbf{Weibo-21}} & \multicolumn{4}{c}{\textbf{GossipCop}} \\
        \cmidrule(lr){2-5} \cmidrule(lr){6-9} \cmidrule(lr){10-13}
        \textbf{Method} & \textbf{Acc} & \textbf{F1-Fake} & \textbf{F1-Real} & \textbf{AUC} & \textbf{Acc} & \textbf{F1-Fake} & \textbf{F1-Real} & \textbf{AUC} & \textbf{Acc} & \textbf{F1-Fake} & \textbf{F1-Real} & \textbf{AUC} \\
        \midrule
        
        \rowcolor{green!20}
        % 3. 缩短标题文字，去掉 "methods" 等冗余词
        \multicolumn{13}{c}{\textit{Multimodal Multi-domain}} \\ 
        \midrule
        EANN & 0.827 & 0.829 & 0.825 & 0.873 & 0.870 & 0.862 & 0.875 & 0.894 & 0.864 & 0.594 & 0.920 & 0.852\\
        FND-CLIP & 0.907 & 0.908 & 0.907 & 0.953 & 0.943 & 0.940 & 0.946 & 0.962 & 0.880 & 0.638 & 0.928 & 0.871\\
        MIMoE-FND & 0.928 & 0.928 & 0.928 & 0.972 & 0.956 & 0.955 & 0.957 & 0.977 & 0.895 & 0.698 & 0.938 & 0.879\\
        KEN & 0.935 & 0.935 & 0.934 & 0.967 & 0.935 & 0.937 & 0.932 & 0.971 & 0.881 & 0.646 & 0.928 & 0.873 \\
        
        \midrule
        
        \rowcolor{orange!20}
        \multicolumn{13}{c}{\textit{Multimodal Single-domain}} \\ 
        \midrule
        CAFE & 0.840 & 0.842 & 0.837 & 0.892 & 0.882 & 0.885 & 0.876 & 0.909 & 0.867 & 0.587 & 0.921 & 0.852 \\
        BMR & 0.918 & 0.914 & 0.904 & 0.954 & 0.929 & 0.927 & 0.925 & 0.962 & 0.895 & 0.691 & 0.876 & 0.881 \\
        RaCMC & 0.915 & 0.917 & 0.914 & 0.921 & 0.942 & 0.938 & 0.943 & 0.962 & 0.879 & 0.641 & 0.927 & 0.838 \\
        SEER & 0.929& 0.928 & 0.939 & 0.934 & 0.932 & 0.927 & 0.925 & 0.960 & 0.893 & 0.673 & 0.871 & 0.875 \\
        
        \midrule
        
        \rowcolor{cyan!20}
        \multicolumn{13}{c}{\textit{LLM-based Reasoning}} \\ 
        \midrule
        INSIDE & 0.881 & 0.684 & 0.932 & 0.910 & 0.896 & 0.816 & 0.832 & 0.871 & 0.900 & 0.707 & 0.934 & 0.881\\
        GLPN-LLM & 0.920 & 0.939 & 0.921 & 0.954 & 0.925 & 0.937 & 0.924 & 0.959 & 0.890 & 0.682 & 0.933 & 0.864\\
        LIFE & 0.924 & 0.940 & 0.921 & 0.958 & 0.928 & 0.935 & 0.934 & 0.975 & 0.864 & 0.701 & 0.866 & 0.855\\
        
        \midrule
        
        \rowcolor{yellow!20}
        \multicolumn{13}{c}{\textbf{Ours}} \\ 
        \midrule
        \textbf{DCCF} & \textcolor{red}{0.951} & \textcolor{red}{0.951} & \textcolor{red}{0.954} & \textcolor{red}{0.978} & \textcolor{red}{0.957} & \textcolor{red}{0.957} & \textcolor{red}{0.965} & \textcolor{red}{0.982} & \textcolor{red}{0.904} & \textcolor{red}{0.723} & \textcolor{red}{0.946} & \textcolor{red}{0.889} \\
        % 移除 Improve 和 p-val 之间的 addlinespace，或者减小它
        Improv. & 1.71\%$\uparrow$ & 1.17\%$\uparrow$ & 1.60\%$\uparrow$ & 0.62\%$\uparrow$ & 0.10\%$\uparrow$ & 0.21\%$\uparrow$ & 0.84\%$\uparrow$ & 0.51\%$\uparrow$ & 0.44\%$\uparrow$ & 2.26\%$\uparrow$ & 0.85\%$\uparrow$ & 0.91\%$\uparrow$ \\
        \bottomrule
    \end{tabular}%
    }
\end{table*}

\begin{enumerate}[label=\arabic*., leftmargin=*]
\item \textbf{Maximally Informative Conflicts.} We identify the pair $(f_i', f_j')$ in $T'$ with maximum tension as the key local conflict $I_{\text{conflict}}$:
\begin{equation}
I_{\text{conflict}} = \text{concat}(f_i', f_j') \quad \text{where } (i, j) = \arg\max(T')
\end{equation}
\item \textbf{Global Consensus.} We compute the mean of $S'$ as $C_{\text{consensus}}$ to represent the global context/tone:
\begin{equation}
C_{\text{consensus}} = \frac{1}{n} \sum_{i=1}^{n} f_i'
\end{equation}
\end{enumerate}

Tone-reference standardization. We concatenate $I_{\text{conflict}}$ and $C_{\text{consensus}}$ into an MLP $g_{std}$. This Tone-reference inconsistency standardization uses consensus to standardize conflict, ensuring that the magnitude of discrepancy is evaluated relative to the document's specific semantic baseline rather than in isolation:
\begin{equation}
V = g_{std}(\text{concat}(I_{\text{conflict}}, C_{\text{consensus}}))
\end{equation}

The outputs are refined fact ($V^F$) and sentiment ($V^E$) inconsistency vectors.

\subsection{Multi-View Deliberative Judgment}
This stage deliberates on the standardized inconsistencies from both semantic subspaces to form a final judgment.

View fusion.
We concatenate $V^F$ and $V^S$ into a final representation $V_{\text{final}}$, encapsulating complementary dual-view inconsistency:
\begin{equation}
V_{\text{final}} = \text{concat}(V^F, V^S)
\end{equation}

Final classification.
$V_{\text{final}}$ is fed into a classifier (e.g., MLP with sigmoid) to predict the probability $\hat{y}$ that the news is Fake.

\subsection{Loss Function}
The total loss $\mathcal{L}_{\text{total}}$ combines the prediction BCE loss $\mathcal{L}_{\text{final}}$ with auxiliary losses $\mathcal{L}_F$ and $\mathcal{L}_E$, balanced by $\lambda_F$ and $\lambda_E$:
\begin{equation}
\mathcal{L}_{\text{total}} = (1 - \lambda_F - \lambda_E)\mathcal{L}_{\text{final}} + \lambda_F \mathcal{L}_F + \lambda_E \mathcal{L}_E
\end{equation}

Joint optimization ensures the model accurately captures visual objective cues, textual sentiment nuances, and critical inconsistencies, yielding robust detection performance.

% 使用 table* 环境跨两栏，以容纳宽表格

\section{Experiments}
\subsection{Experimental Settings}
We validate DCCF with experiments detailing our datasets, baselines, and implementation.

\textbf{Datasets.}
We use three benchmarks: Weibo \cite{wang2018eann}, Weibo21 \cite{zhou2020safe}, and GossipCop \cite{liu2025modality}, following established protocols. Weibo \cite{wang2018eann} includes 7,532 training (3,749 real/3,783 fake) and 1,996 test (996 real/1,000 fake) articles. Weibo21 \cite{zhou2020safe} has 9,127 total articles (4,640 real/4,487 fake). GossipCop \cite{liu2025modality} provides 10,010 training (7,974 real/2,036 fake) and 2,830 testing (2,285 real/545 fake) instances.

% 建议：由于表格列数较多，改为跨栏表格 (table*) 以保证字体大小可读
\begin{table*}[t]
\centering
\caption{Ablation study results showing performance drops for different model variants.}
\label{tab:ablation_moda_llm}
\resizebox{0.9\textwidth}{!}{% 使用 textwidth 适应跨栏宽度
\begin{tabular}{lccccccccc}
\toprule
\multirow{2}{*}{Model} & \multicolumn{3}{c}{\textbf{Weibo}} & \multicolumn{3}{c}{\textbf{Weibo-21}} & \multicolumn{3}{c}{\textbf{GossipCop}} \\
\cmidrule(lr){2-4} \cmidrule(lr){5-7} \cmidrule(lr){8-10}
& \textbf{Acc} & \textbf{F1-Fake} & \textbf{F1-Real} & \textbf{Acc} & \textbf{F1-Fake} & \textbf{F1-Real} & \textbf{Acc} & \textbf{F1-Fake} & \textbf{F1-Real} \\
\midrule
\textbf{DCCF} & 0.951 & 0.951 & 0.954 & 0.957 & 0.957 & 0.965 & 0.904 & 0.723 & 0.946 \\
\midrule
\rowcolor{green!20}
\multicolumn{10}{c}{\textit{Multi-task Feature Learning}} \\
\midrule
- w/o $\mathcal{L}_{\text{Yolo}}$ & 0.925\,\textcolor{red!80!black}{\tiny{$\downarrow$2.7\%}} & 0.926\,\textcolor{red!80!black}{\tiny{$\downarrow$2.6\%}} & 0.922\,\textcolor{red!80!black}{\tiny{$\downarrow$3.4\%}} & 0.933\,\textcolor{red!80!black}{\tiny{$\downarrow$2.5\%}} & 0.930\,\textcolor{red!80!black}{\tiny{$\downarrow$2.8\%}} & 0.934\,\textcolor{red!80!black}{\tiny{$\downarrow$3.2\%}} & 0.885\,\textcolor{red!80!black}{\tiny{$\downarrow$2.1\%}} & 0.679\,\textcolor{red!80!black}{\tiny{$\downarrow$6.1\%}} & 0.922\,\textcolor{red!80!black}{\tiny{$\downarrow$2.5\%}} \\
- w/o $\mathcal{L}_{\text{SenticNet}}$ & 0.927\,\textcolor{red!80!black}{\tiny{$\downarrow$2.5\%}} & 0.920\,\textcolor{red!80!black}{\tiny{$\downarrow$3.3\%}} & 0.927\,\textcolor{red!80!black}{\tiny{$\downarrow$2.8\%}} & 0.928\,\textcolor{red!80!black}{\tiny{$\downarrow$3.0\%}} & 0.932\,\textcolor{red!80!black}{\tiny{$\downarrow$2.6\%}} & 0.939\,\textcolor{red!80!black}{\tiny{$\downarrow$2.7\%}} & 0.880\,\textcolor{red!80!black}{\tiny{$\downarrow$2.7\%}} & 0.678\,\textcolor{red!80!black}{\tiny{$\downarrow$6.2\%}} & 0.919\,\textcolor{red!80!black}{\tiny{$\downarrow$2.9\%}} \\
- w/o $\mathcal{L}_{\text{Both}}$ & 0.922\,\textcolor{red!80!black}{\tiny{$\downarrow$3.1\%}} & 0.912\,\textcolor{red!80!black}{\tiny{$\downarrow$4.1\%}} & 0.908\,\textcolor{red!80!black}{\tiny{$\downarrow$4.8\%}} & 0.941\,\textcolor{red!80!black}{\tiny{$\downarrow$1.7\%}} & 0.917\,\textcolor{red!80!black}{\tiny{$\downarrow$4.2\%}} & 0.920\,\textcolor{red!80!black}{\tiny{$\downarrow$4.7\%}} & 0.885\,\textcolor{red!80!black}{\tiny{$\downarrow$2.1\%}} & 0.673\,\textcolor{red!80!black}{\tiny{$\downarrow$6.9\%}} & 0.909\,\textcolor{red!80!black}{\tiny{$\downarrow$3.9\%}} \\
\midrule
\rowcolor{orange!20}
\multicolumn{10}{c}{\textit{DARFU}} \\
\midrule
- w/o \text{Evolution} & 0.931\,\textcolor{red!80!black}{\tiny{$\downarrow$2.1\%}} & 0.928\,\textcolor{red!80!black}{\tiny{$\downarrow$2.4\%}} & 0.924\,\textcolor{red!80!black}{\tiny{$\downarrow$3.1\%}} & 0.943\,\textcolor{red!80!black}{\tiny{$\downarrow$1.5\%}} & 0.936\,\textcolor{red!80!black}{\tiny{$\downarrow$2.2\%}} & 0.939\,\textcolor{red!80!black}{\tiny{$\downarrow$2.7\%}} & 0.879\,\textcolor{red!80!black}{\tiny{$\downarrow$2.8\%}} & 0.692\,\textcolor{red!80!black}{\tiny{$\downarrow$4.3\%}} & 0.928\,\textcolor{red!80!black}{\tiny{$\downarrow$1.9\%}} \\
- w/o \text{Tension Weighting} & 0.934\,\textcolor{red!80!black}{\tiny{$\downarrow$1.8\%}} & 0.924\,\textcolor{red!80!black}{\tiny{$\downarrow$2.8\%}} & 0.922\,\textcolor{red!80!black}{\tiny{$\downarrow$3.4\%}} & 0.947\,\textcolor{red!80!black}{\tiny{$\downarrow$1.0\%}} & 0.934\,\textcolor{red!80!black}{\tiny{$\downarrow$2.4\%}} & 0.935\,\textcolor{red!80!black}{\tiny{$\downarrow$3.1\%}} & 0.870\,\textcolor{red!80!black}{\tiny{$\downarrow$3.8\%}} & 0.686\,\textcolor{red!80!black}{\tiny{$\downarrow$5.1\%}} & 0.922\,\textcolor{red!80!black}{\tiny{$\downarrow$2.5\%}} \\
\midrule
\rowcolor{cyan!20}
\multicolumn{10}{c}{\textit{Conflict-Consensus Metric Extraction}} \\
\midrule
- w/o \text{Maximally Informative Conflicts} & 0.915\,\textcolor{red!80!black}{\tiny{$\downarrow$3.8\%}} & 0.918\,\textcolor{red!80!black}{\tiny{$\downarrow$3.5\%}} & 0.920\,\textcolor{red!80!black}{\tiny{$\downarrow$3.6\%}} & 0.911\,\textcolor{red!80!black}{\tiny{$\downarrow$4.8\%}} & 0.923\,\textcolor{red!80!black}{\tiny{$\downarrow$3.6\%}} & 0.924\,\textcolor{red!80!black}{\tiny{$\downarrow$4.3\%}} & 0.863\,\textcolor{red!80!black}{\tiny{$\downarrow$4.5\%}} & 0.683\,\textcolor{red!80!black}{\tiny{$\downarrow$5.5\%}} & 0.921\,\textcolor{red!80!black}{\tiny{$\downarrow$2.6\%}} \\
- w/o \text{Global Consensus} & 0.900\,\textcolor{red!80!black}{\tiny{$\downarrow$5.4\%}} & 0.912\,\textcolor{red!80!black}{\tiny{$\downarrow$4.1\%}} & 0.913\,\textcolor{red!80!black}{\tiny{$\downarrow$4.3\%}} & 0.919\,\textcolor{red!80!black}{\tiny{$\downarrow$4.0\%}} & 0.927\,\textcolor{red!80!black}{\tiny{$\downarrow$3.1\%}} & 0.932\,\textcolor{red!80!black}{\tiny{$\downarrow$3.4\%}} & 0.858\,\textcolor{red!80!black}{\tiny{$\downarrow$5.1\%}} & 0.677\,\textcolor{red!80!black}{\tiny{$\downarrow$6.4\%}} & 0.922\,\textcolor{red!80!black}{\tiny{$\downarrow$2.5\%}} \\
\midrule
\rowcolor{yellow!20}
\multicolumn{10}{c}{\textit{Multi-View Deliberative Judgment}} \\
\midrule
- w/o \text{Fact View} & 0.913\,\textcolor{red!80!black}{\tiny{$\downarrow$4.0\%}} & 0.918\,\textcolor{red!80!black}{\tiny{$\downarrow$3.5\%}} & 0.922\,\textcolor{red!80!black}{\tiny{$\downarrow$3.4\%}} & 0.920\,\textcolor{red!80!black}{\tiny{$\downarrow$3.9\%}} & 0.925\,\textcolor{red!80!black}{\tiny{$\downarrow$3.3\%}} & 0.930\,\textcolor{red!80!black}{\tiny{$\downarrow$3.6\%}} & 0.861\,\textcolor{red!80!black}{\tiny{$\downarrow$4.8\%}} & 0.671\,\textcolor{red!80!black}{\tiny{$\downarrow$7.2\%}} & 0.908\,\textcolor{red!80!black}{\tiny{$\downarrow$4.0\%}} \\
- w/o \text{Sentiment View} & 0.919\,\textcolor{red!80!black}{\tiny{$\downarrow$3.4\%}} & 0.916\,\textcolor{red!80!black}{\tiny{$\downarrow$3.7\%}} & 0.924\,\textcolor{red!80!black}{\tiny{$\downarrow$3.1\%}} & 0.933\,\textcolor{red!80!black}{\tiny{$\downarrow$2.5\%}} & 0.926\,\textcolor{red!80!black}{\tiny{$\downarrow$3.2\%}} & 0.931\,\textcolor{red!80!black}{\tiny{$\downarrow$3.5\%}} & 0.887\,\textcolor{red!80!black}{\tiny{$\downarrow$1.9\%}} & 0.684\,\textcolor{red!80!black}{\tiny{$\downarrow$5.4\%}} & 0.926\,\textcolor{red!80!black}{\tiny{$\downarrow$2.1\%}} \\
\bottomrule
\end{tabular}
}%
\end{table*}

\textbf{Baselines.}
We benchmark against three categories: (1) Unimodal methods (MVAN \cite{ni2021mvan}, SpotFake \cite{singhal2019spotfake}). (2) Cross-domain generalization (EANN \cite{wang2018eann}, FND-CLIP \cite{zhou2023multimodal}, MIMoE-FND \cite{liu2025modality}, KEN \cite{zhu2025ken}). (3) LLM distillation (GLPN-LLM \cite{hu2025synergizing}, INSIDE \cite{wang2025bridging}, LIFE \cite{wang2025prompt}).

\textbf{Implementation Details.}
Visual features used a pretrained MAE \cite{he2022masked} with $224\times224$ images. Text used bert-base-chinese \cite{devlin2019bert} (Weibo/Weibo21) and bert-base-uncased \cite{devlin2019bert} (GossipCop), truncated to 197 tokens. Features were aligned using CLIP \cite{yang2022chinese}. The auxiliary loss coefficients $\lambda_F$ and $\lambda_E$ were both set to 0.075. The model used PyTorch, trained on one NVIDIA RTX 4090 GPU for 50 epochs with early stopping.

\subsection{Overall Performance}
To validate DCCF's superiority, we compare it against 11 baselines on three datasets (Table \ref{tab:main_results}). From the results, we draw these key observations:

\textbf{(O1): DCCF consistently achieves state-of-the-art performance} across diverse benchmarks. On Weibo, it surpasses the strongest baseline (LIFE) by 1.1\% in accuracy and 2.7\% in F1-Fake, highlighting its capability in detecting standard multimodal inconsistencies.

\textbf{(O2): On the recent Weibo-21 dataset, DCCF maintains a competitive edge.} Despite a narrower margin against the top baseline (MIMOE-FND), our model secures the best results across all four metrics. This confirms that our dynamic conflict-seeking paradigm generalizes effectively to varying data distributions.

\textbf{(O3): DCCF demonstrates superior robustness on the imbalanced GossipCop} (80\% real). Unlike baselines that trade off F1 scores, DCCF achieves the highest accuracy (0.904) and balanced performance (F1-Fake: 0.723, F1-Real: 0.946), effectively mitigating class imbalance pitfalls.

\textbf{(O4): The results validate our architectural hypothesis.} Disentangling features into fact/sentiment spaces and amplifying inconsistency via the tension field network proves more effective than both traditional fusion (e.g., BMR \cite{ying2023bootstrapping}, SEER \cite{zhu2025seer}) and LLM-distillation approaches (e.g., INSIDE \cite{wang2025bridging}, LIFE \cite{wang2025prompt}).

\subsection{Ablation Study}
To understand DCCF's core components, we ran an ablation study (Table \ref{tab:ablation_moda_llm}).

Effect of Multi-View Deliberative Judgment. Removing the Fact View (w/o Fact View) or Sentiment View (w/o Sentiment View) degraded performance, especially the former. This confirms integrating inconsistency signals from factual and sentimental spaces, validating our multi-view architecture.

Effect of Feature Disentanglement Components. Removing auxiliary losses (w/o $\mathcal{L}_{\text{Yolo}}$) or (w/o $\mathcal{L}_{\text{SenticNet}}$) hurt performance, especially $\mathcal{L}_{\text{Yolo}}$, verifying the importance of multi-task learning. Removing both (w/o $\mathcal{L}_{\text{Double}}$) caused a severe drop, proving this separation is critical.

\begin{figure}[t]
    \centering
    % --- 第一张图（上） ---
    % 修改点1：宽度改为 1.0\linewidth (占满整行)
    \begin{subfigure}{0.9\linewidth}
        \centering
        \includegraphics[width=\linewidth]{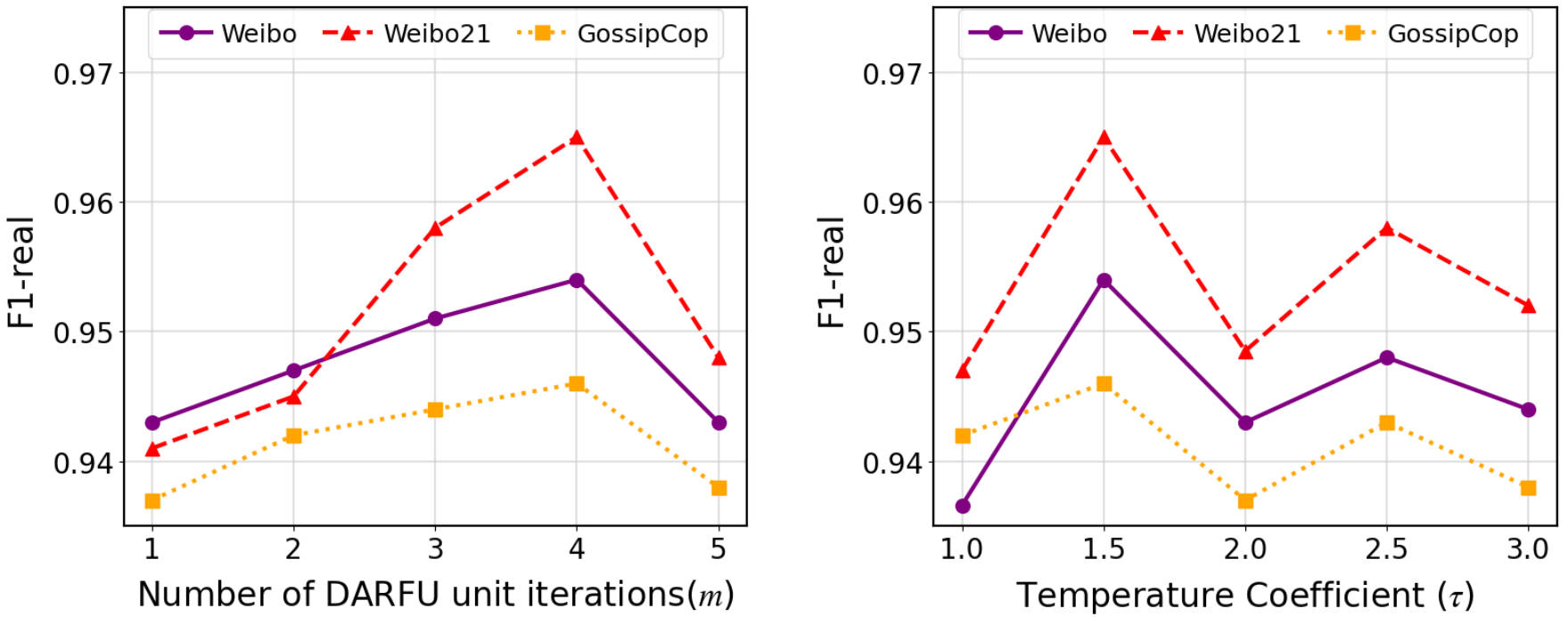}
        \caption{Effects of the number of DARFU unit iterations (m) and the temperature coefficient ($\tau$).}
        \label{fig:hyper_a}
    \end{subfigure}
    
    % 修改点2：添加垂直间距，迫使换行
    \par\bigskip % 或者使用 \vspace{5pt} 来精确控制间距
    
    % --- 第二张图（下） ---
    \begin{subfigure}{0.9\linewidth}
        \centering
        \includegraphics[width=\linewidth]{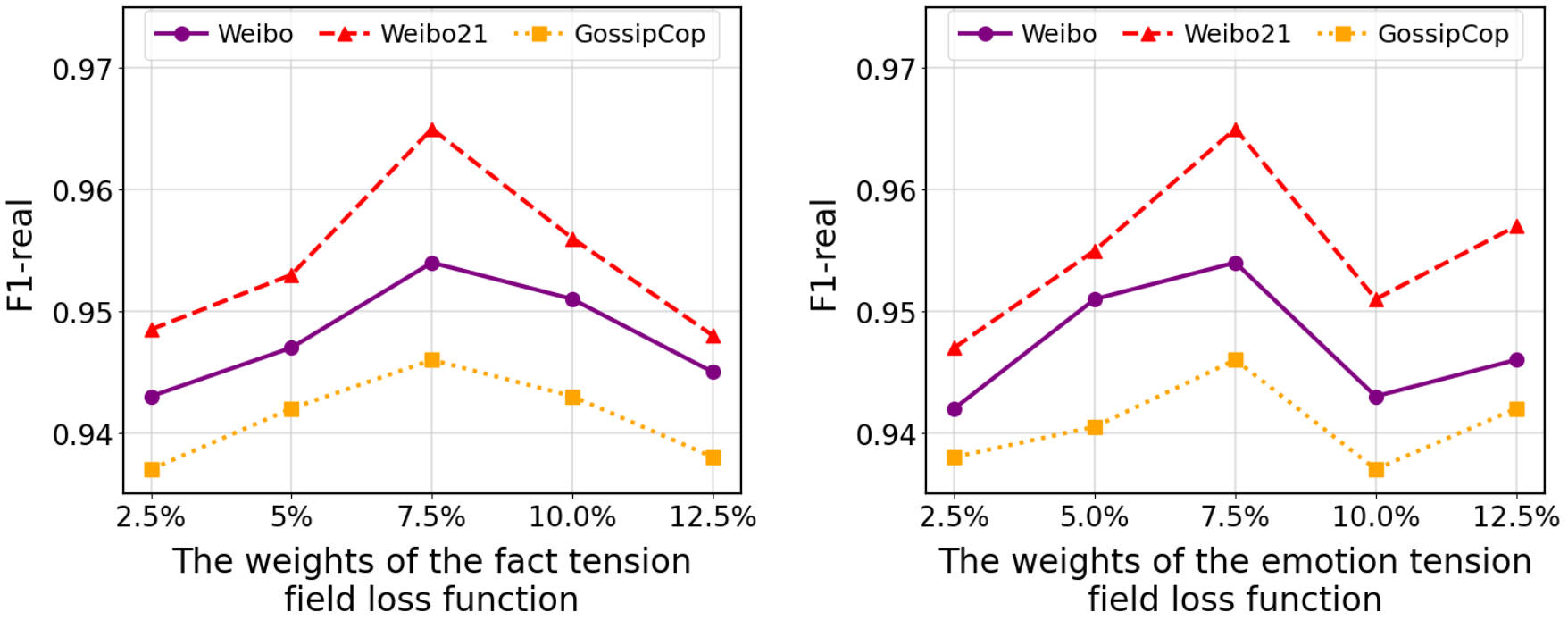}
        \caption{Effects of the fact auxiliary loss coefficient ($\mathcal{L}_F$) and the sentiment auxiliary loss coefficient ($\mathcal{L}_E$).}
        \label{fig:hyper_b}
    \end{subfigure}
    
    \caption{Analysis of hyperparameter sensitivity. This figure shows the impact of four different hyperparameters on the model's F1-real score across three datasets. (a) Effects of the number of DARFU unit iterations (m) and the Temperature Coefficient ($\tau$). (b) Effects of the fact auxiliary loss coefficient ($\mathcal{L}_F$) and the sentiment auxiliary loss coefficient ($\mathcal{L}_E$).}
    \label{fig:hyperparameter_sensitivity}
    
    \vspace{-18pt}
\end{figure}

Effect of Tension Field Network and Metric Extraction. Removing the Feature Dynamics Evolution stage (w/o Evolution) caused a severe performance drop (e.g., 5.2\% on Weibo). Removing the tension-to-weight mechanism (w/o Tension Weighting) also degraded performance. Furthermore, removing metric components (w/o Global Consensus or w/o Maximally Informative Conflicts) caused degradation, demonstrating the effectiveness of DCCF components.

\subsection{Parameter Sensitivity Analysis}
We analyzed parameter sensitivity on three datasets, focusing on four key parameters: DARFU iterations (m), temperature coefficient ($\tau$), and the fact auxiliary loss coefficient ($\lambda_F$) and the sentiment auxiliary loss coefficient ($\lambda_E$). As shown in the figures, the model is robust. Performance forms a bell shaped curve, peaking at m=4, $\tau$=1.5, and a 7.5\% loss weight for both fields. Performance gracefully declines from these optimal points but remains high, demonstrating the model is effective across various configurations and not overly sensitive.

\begin{table}[t]
\centering
\caption{DCCF explainability case study on two challenging examples from the GossipCop test set. The calibration allocation abbreviations denote: FT (Fact-Text), FI (Fact-Image), ST (Sentiment-Text), and SI (Sentiment-Image).}
\label{tab:adaptive_structure_transposed}
\resizebox{\linewidth}{!}{%
\renewcommand{\arraystretch}{1.5}%
\begin{tabular}{>{\bfseries}l c c c c c c}
\toprule
& \textbf{\makecell{News\\Posts}} 
& \textbf{\makecell{False\\Type}} 
& \textbf{\makecell{Ground\\Truth}} 
& \textbf{\makecell{Calibration\\Allocation}} 
& \textbf{\makecell{DCCF}} 
& \textbf{\makecell{MIMoE-FND}} \\
\midrule
% --- Case 1 ---
\raisebox{4\height}{Case 1} &
\includegraphics[width=0.2\textwidth]{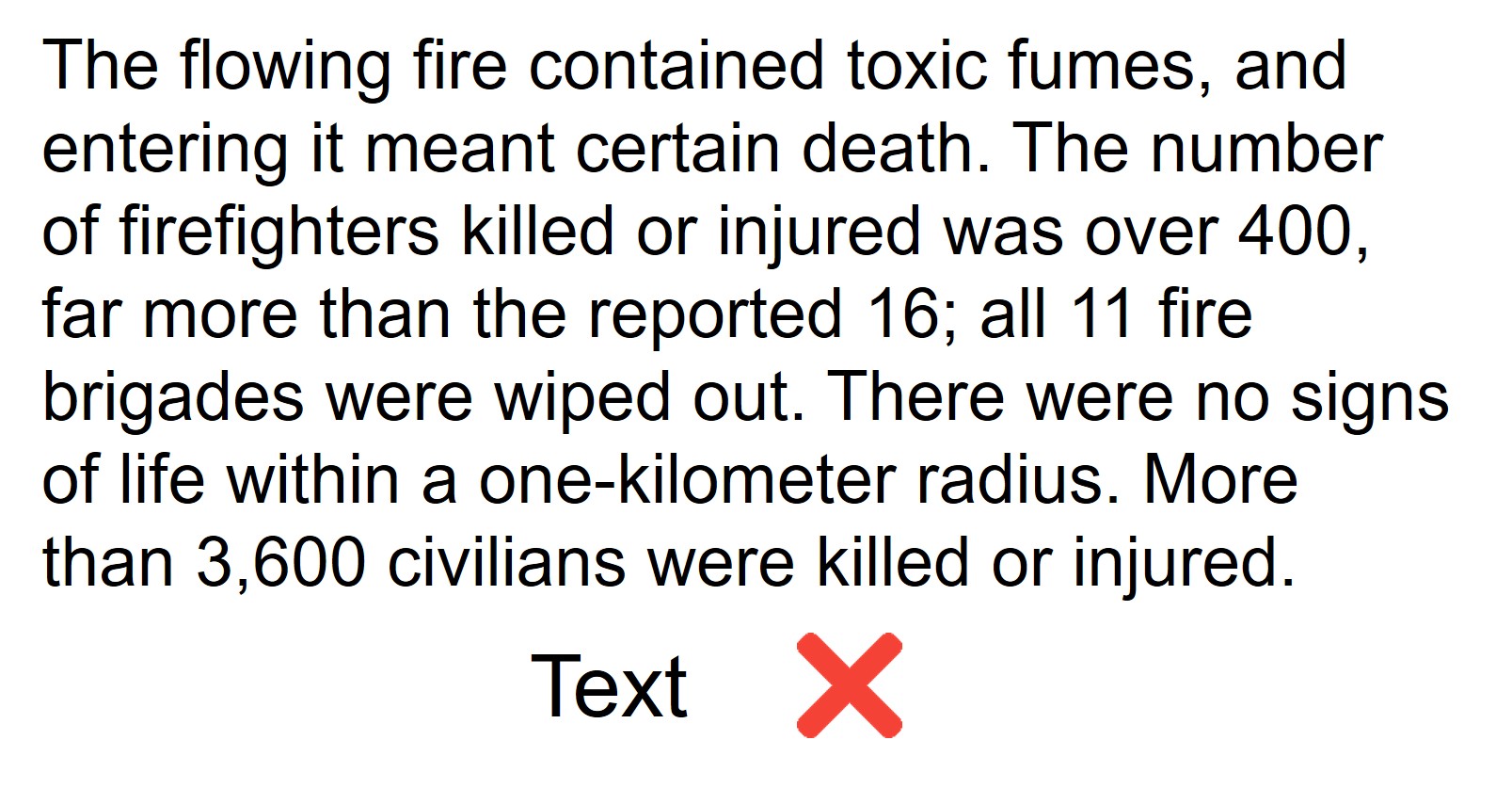} &
\raisebox{2\height}{\makecell{Text \\ Fabrication}}&
\raisebox{2\height}{\makecell{Fake \\ news}} &
\includegraphics[width=0.2\textwidth]{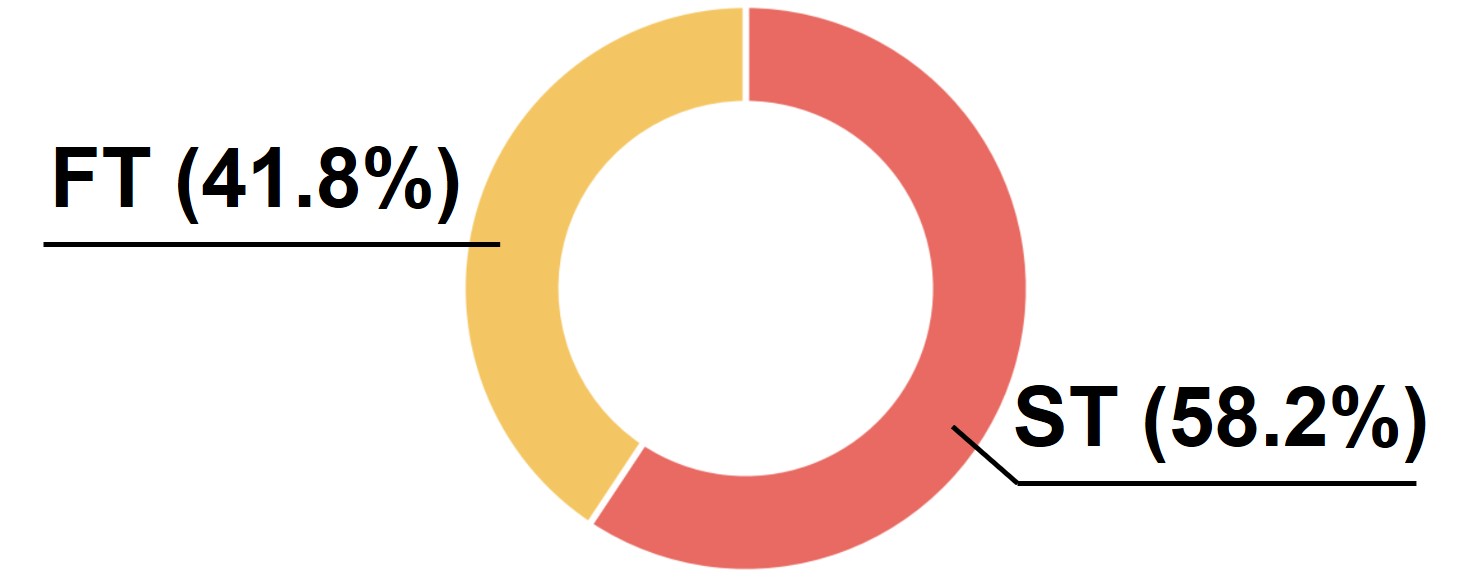} &
\raisebox{2\height}{\makecell{Fake \\ news}}  &
\raisebox{2\height}{\makecell{Fake \\ news}}  \\
\midrule
% --- Case 2 ---
\raisebox{4\height}{Case 2} &
\includegraphics[width=0.2\textwidth]{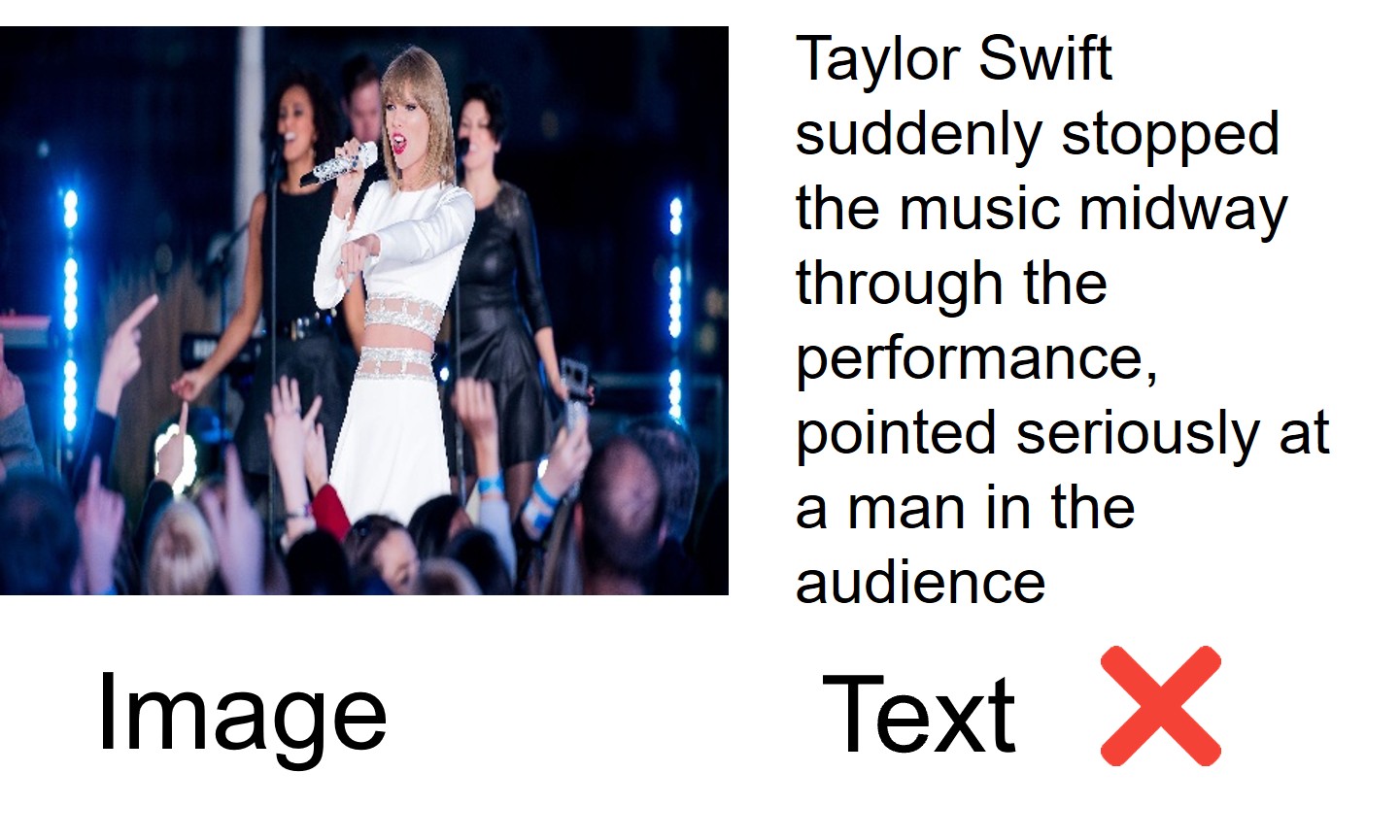} &
\raisebox{2\height}{\makecell{Image-Text \\ Mismatch}} &
\raisebox{2\height}{\makecell{Fake \\ news}} &
\includegraphics[width=0.2\textwidth]{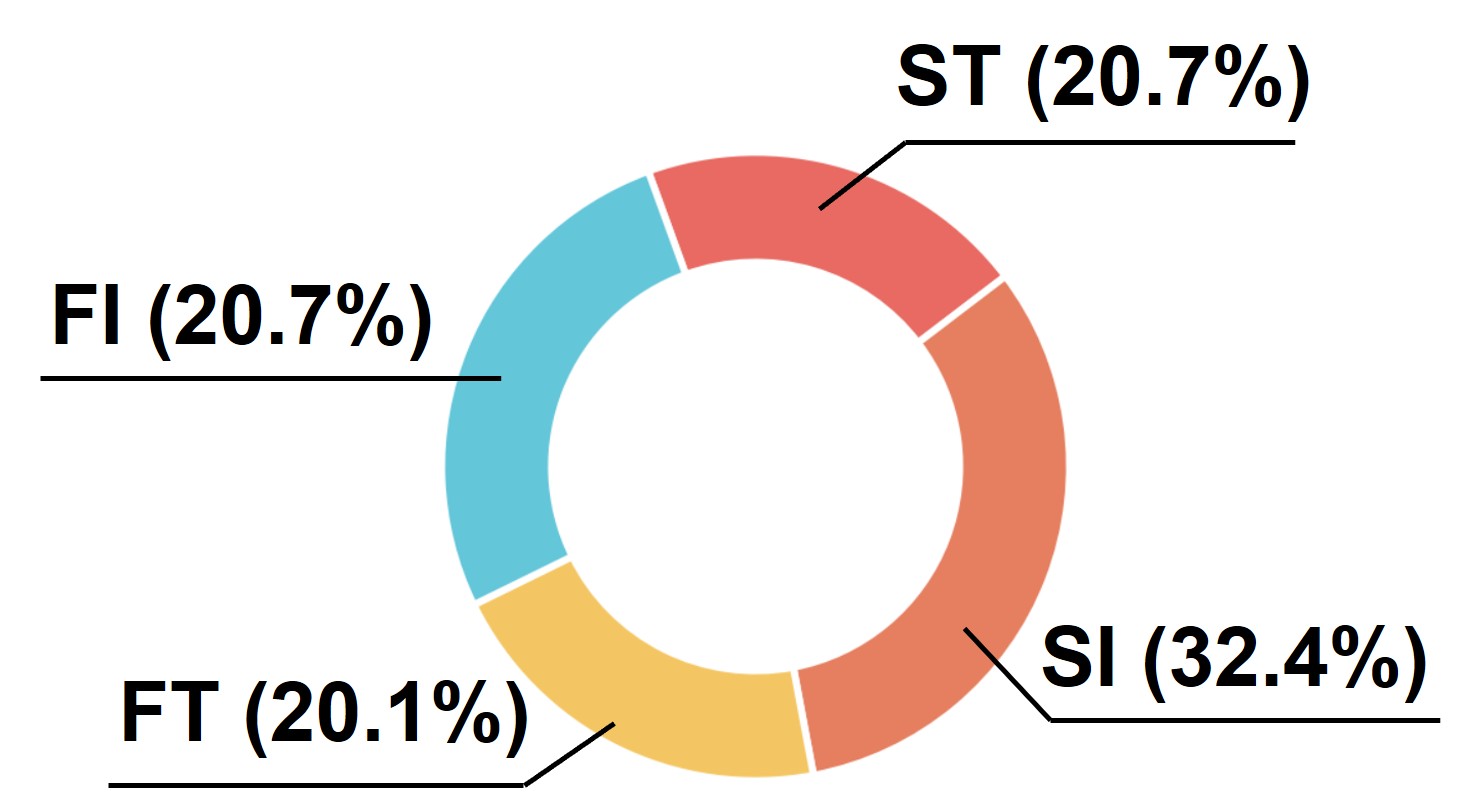} &
\raisebox{2\height}{\makecell{Fake \\ news}} &
\raisebox{2\height}{\makecell{Real \\ news}}  \\
\bottomrule
\end{tabular}
}
\vspace{-9pt}
\end{table}

\subsection{Case Study: Model Explainability}
We investigate DCCF's interpretability with a case study of two challenging GossipCop instances. This illustrates how DCCF acquires and distills deep reasoning into its text and image representations.

The first case is a Text Fabrication. Both DCCF and the baseline MIMoE-FND \cite{liu2025modality} correctly classified it.

The second case is an Image-Text mismatch. DCCF correctly predicted Fake news, while the baseline MIMoE-FND \cite{liu2025modality} failed, misclassifying it as Real news.

\subsection{T-SNE Visualizations}
\begin{figure}[t]
\centering
\includegraphics[width=0.8\columnwidth]{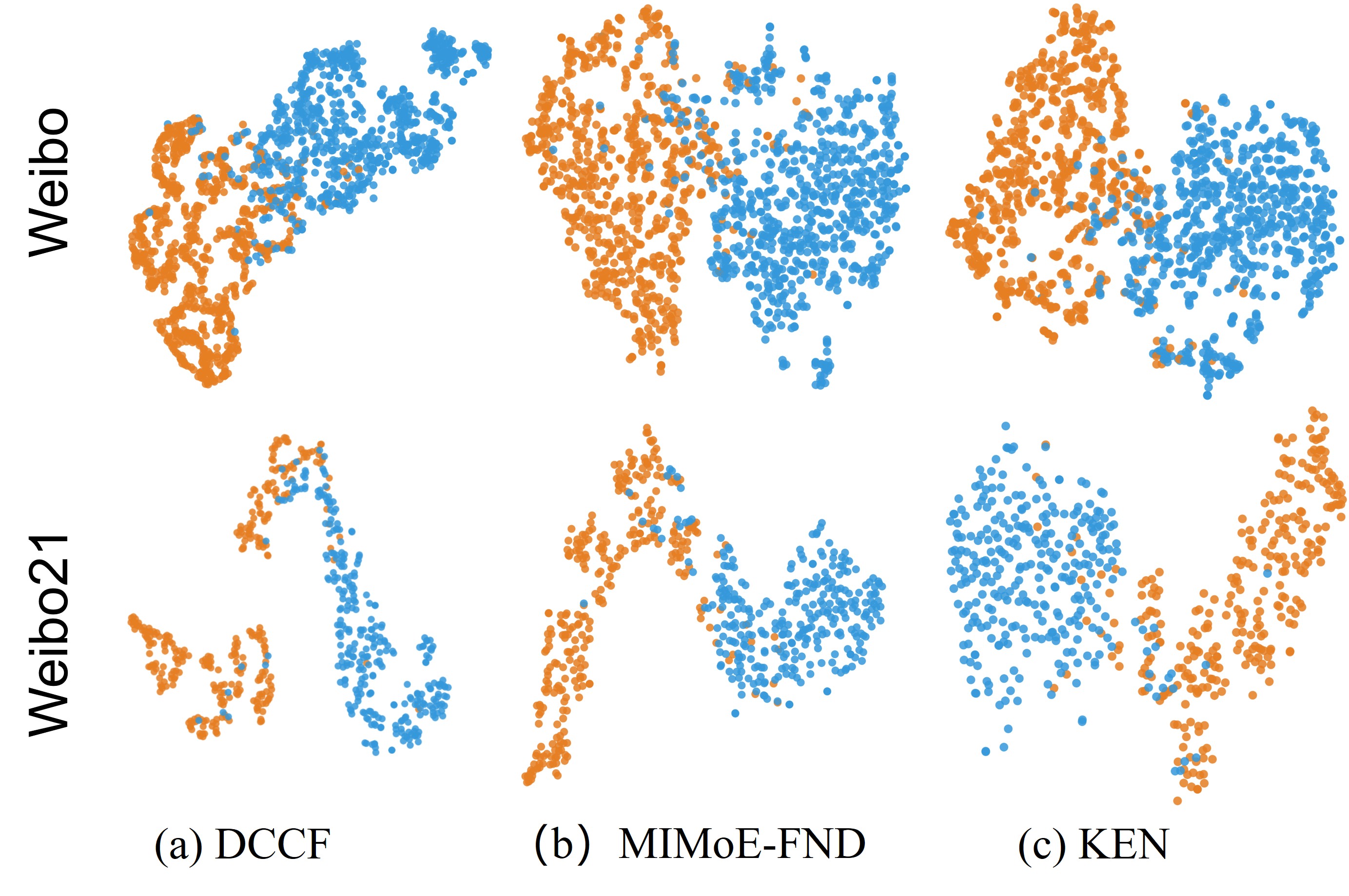} % 调整宽度比例以适应单栏
\caption{T-SNE visualization of test set features. Same color dots indicate the same label.}
\label{fig:tsne_visualization}
\vspace{-9pt}
\end{figure}

Fig. \ref{fig:tsne_visualization} shows T-SNE visualizations of features from DCCF, MIMoE-FND \cite{liu2025modality}, and KEN \cite{zhu2025ken} on the Weibo and Weibo21 test sets. Compared to the baselines, DCCF produces fewer fake news outliers and less overlap between real and fake news embeddings, confirming its superior performance.

On Weibo21, DCCF's features form multiple, clearly separated subclusters, unlike the single clusters on Weibo. This suggests Weibo21 has varying topics and that DCCF not only distinguishes authenticity but also captures deep, event-level semantic information, spatially distinguishing different events.

\section{Conclusion}
In this paper, we propose the DCCF, a novel inconsistency-seeking paradigm. DCCF initially employs fact sentiment feature extraction guided by multi-task supervision to decouple semantic spaces. Subsequently, the fact sentiment tension field network iteratively models feature dynamics to polarize representations, distilling interpretable maximally informative conflicts and global consensus metrics. Finally, the Multi-View Deliberative Judgment fuses these standardized indicators for robust detection. Extensive experiments validate DCCF's superiority. A primary limitation is the framework's reliance on the quality of auxiliary pseudo-labels, where upstream noise may propagate to the decoupled spaces. Future work will explore integrating Large Language Models to enhance the robustness of these semantic constraints.

\bibliographystyle{IEEEbib}
\bibliography{icme2026references}

\end{document}